%% file: root_IEEEtrans.tex
\documentclass[conference,final]{IEEEtran}
\IEEEoverridecommandlockouts

\usepackage{cite}
\usepackage{amsmath,amssymb,amsfonts}
\usepackage{algorithmic}
\usepackage{graphicx}
\usepackage{textcomp}
\usepackage{xcolor}

\usepackage{lipsum}
\usepackage{subcaption}
\usepackage{array}
\usepackage{booktabs}
\usepackage{hyperref}
\usepackage{float}
\usepackage{tikz}

\usepackage{draftwatermark}

\SetWatermarkText{DRAFT}
\SetWatermarkScale{3}  
\SetWatermarkColor[gray]{0.88}  

\def\BibTeX{{\rm B\kern-.05em{\sc i\kern-.025em b}\kern-.08em
    T\kern-.1667em\lower.7ex\hbox{E}\kern-.125emX}}

\newcommand\copyrighttext{%
  \footnotesize \textcopyright \the\year{} IEEE. Personal use of this material is permitted. Permission from IEEE must be obtained for all other uses, including reprinting/republishing this material for advertising or promotional purposes, collecting new collected works for resale or redistribution to servers or lists, or reuse of any copyrighted component of this work in other works. This work is accepted at IEEE/ASME International Conference on Advanced Intelligent Mechatronics (AIM2026).}

\newcommand\copyrightnotice{%
\begin{tikzpicture}[remember picture,overlay]
\node[anchor=south,yshift=10pt] at (current page.south) {\fbox{\parbox{\dimexpr0.75\textwidth-\fboxsep-\fboxrule\relax}{\copyrighttext}}};
\end{tikzpicture}%
}

\begin{document}

\title{Articulated Humanoid Head for a Robot Receptionist Capable of Natural Human Interaction\\
}

\author{\IEEEauthorblockN{ Tharusha Fonseka}
\IEEEauthorblockA{\textit{Dept. of ECE} \\
\textit{National University of Singapore}\\
Singapore \\
tharusha.fonseka@u.nus.edu}
\and
\IEEEauthorblockN{Charuka Bandara}
\IEEEauthorblockA{\textit{Dept. of ENTC} \\
\textit{University of Moratuwa}\\
Moratuwa, Sri Lanka \\
Bandarahmcnk.20@uom.lk}
\and
\IEEEauthorblockN{Moshintha Hewavitharana}
\IEEEauthorblockA{\textit{Dept. of ENTC} \\
\textit{University of Moratuwa}\\
Moratuwa, Sri Lanka \\
hewavitharanami.20@uom.lk}
\and
\IEEEauthorblockN{Melisa Arukgoda}
\IEEEauthorblockA{\textit{Dept. of ENTC} \\
\textit{University of Moratuwa}\\
Moratuwa, Sri Lanka \\
arukgodaamo.20@uom.lk}
\and
\IEEEauthorblockN{Wageesha~N.~ Manamperi}
\IEEEauthorblockA{\textit{Dept. of ENTC} \\
\textit{University of Moratuwa}\\
Moratuwa, Sri Lanka \\
wageesham@uom.lk}
\and
\IEEEauthorblockN{Udaya~S.~K.~Perera~Miriya~Thanthrige}
\IEEEauthorblockA{\textit{Dept. of ENTC} \\
\textit{University of Moratuwa}\\
Moratuwa, Sri Lanka \\
sampathk@uom.lk}
\and
\IEEEauthorblockN{Peshala Jayasekara}
\IEEEauthorblockA{\textit{Dept. of ENTC} \\
\textit{University of Moratuwa}\\
Moratuwa, Sri Lanka \\
peshala@uom.lk}
}

\maketitle
\copyrightnotice
\begin{abstract}
\input{sections/abstract}
\end{abstract}

\begin{IEEEkeywords}
Humanoid Robots, Emotion Expression, Conversational AI, Human Re-identification, Human-Robot Interaction
\end{IEEEkeywords}

\section{INTRODUCTION}
\input{sections/Introduction}

\section{Related Work}
\input{sections/Related_work}

\section{Methodology}
\input{sections/Methodology}


\section{Results and Discussion}
\input{sections/Results}


\section{Conclusion}
\input{sections/Conclusion}

\bibliographystyle{IEEEtran}
\bibliography{ref}

\end{document}

%% file: sections/abstract.tex
Humanoid robots have become increasingly popular in applications such as social interaction, education, and service roles, which drives the need for more natural and efficient human-robot interactions. However, currently available humanoid heads often face limitations, including high costs, mechanical complexity, and limited adaptability across diverse environments. To address these challenges, we present an articulated humanoid robot head designed for a receptionist role, integrating a mechanical structure with 21 degrees of freedom (DoF), including mechanisms for the mouth, eyes, eyebrows, and neck, and covered with realistic silicone skin to achieve a human-like appearance and expression. The system integrates a model-based architecture that combines SCRFD, ArcFace, and ByTetrack for face recognition and Llama and Whisper for natural language processing, with hardware support enabling real-time operations and human re-identification. The conversational ability and re-identification capabilities of the humanoid robot head were quantitatively measured, while its emotional expressiveness and human likeness were evaluated through a user study, achieving an average human likeness score of 4.13 out of 5.

%% file: sections/Introduction.tex
Humanoid robots are increasingly used in social roles such as receptionists. In these roles, the robot’s head is central to human–robot interaction, enabling facial expressions, eye contact, and brief conversations that establish social presence. However, most expressive humanoid heads rely on complex mechanisms with many degrees of freedom (DoFs), making them difficult to maintain for scalable deployment. While robot receptionists do require expressive capability, they only need short, focused interactions rather than deep emotional exchanges. This creates a clear design opportunity to build a mechanically simple, deployable humanoid head that delivers essential expressiveness and interactivity.

Advanced humanoid robots such as Ameca and Mesmer from Engineered Arts~\cite{EngineeredArts2025}, and Aria from Realbotix~\cite{Realbotix2025}, demonstrate high levels of realism and facial expressiveness. The Desktop version of Ameca has 32-DoF solely for facial and neck movements. However, their complexity and cost associated with it make them impractical for widespread deployment in service-specific applications, such as reception. In contrast, simpler designs like EVA~\cite{EVA} and EMO~\cite{EMO} achieve lower mechanical overhead at the cost of reduced naturalness and limited interactive features. EVA lacks both gaze control and verbal interaction, while EMO supports basic gaze control but does not offer conversation capabilities. Most existing designs either over-engineer capabilities not needed for short social exchanges or compromise too much on interaction quality.
This paper presents a humanoid robot head that enables expressive interaction and natural conversation for receptionist applications.

Our contribution in this paper focuses on three aspects:
\begin{enumerate}
    \item A simplified mechanical design with a reduced number of DoFs capable of expressing six basic emotions;
    \item A multi-user interaction system enabling human re-identification and natural conversation in multi-speaker settings, deployed on an edge device;
    \item The public release of the robot design files and source code to support reproducibility. The resources are available at \textit{\url{https://tham-cham.github.io/}}.
\end{enumerate}

%% file: sections/Related_work.tex
\begin{figure*}[!t]
\vspace{2pt}
    \centering
    \includegraphics[width=1\linewidth]{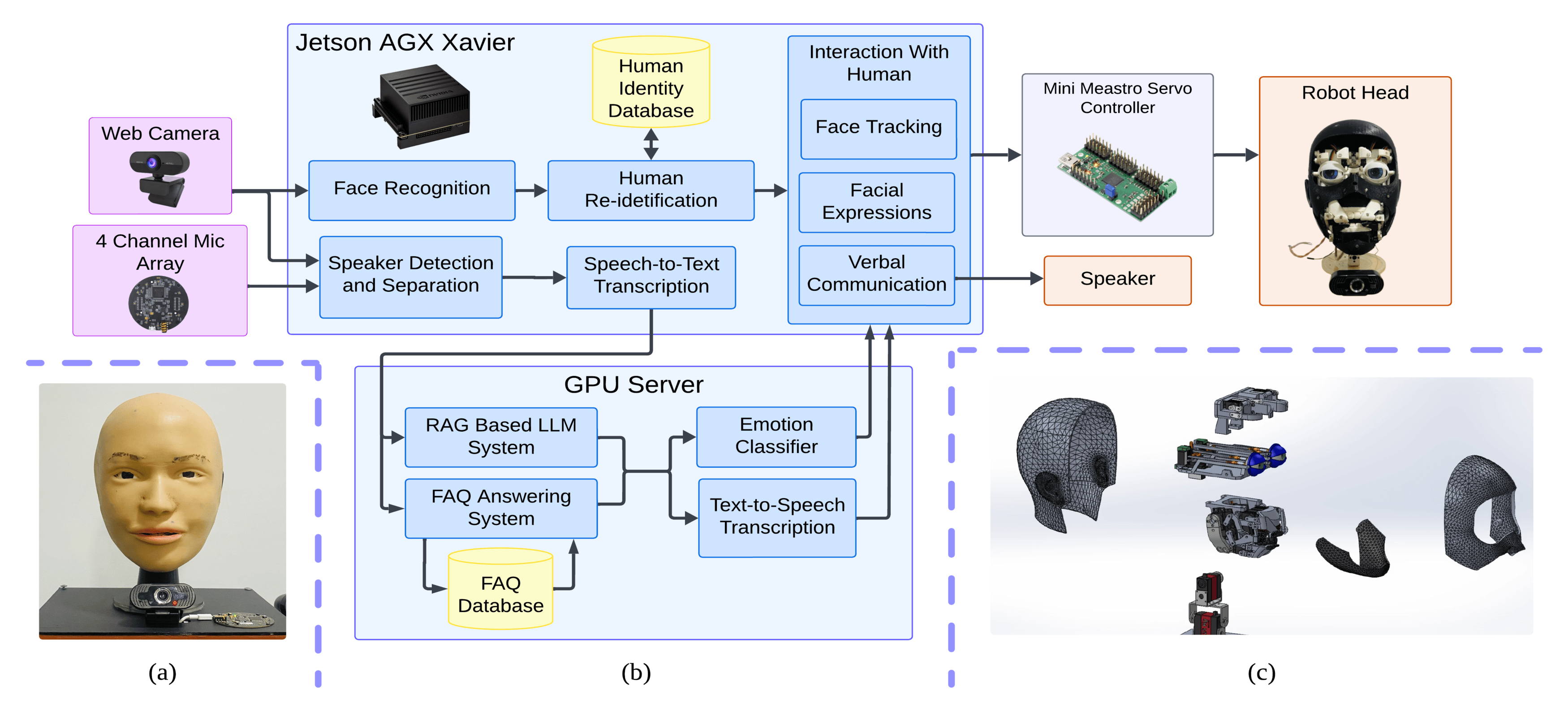}
    \caption{(a) Final appearance of the humanoid head with silicone mask, (b) Overall system architecture: A web camera and a Respeaker mic array provide the input to the main processing unit, Nvidia Jetson AGX Xavier, where the human re-identification process, audio processing, and human interaction tasks are carried out. The external GPU server generates a relevant response. (c) Exploded view of the mechanical design with four submodules and rigid supports for the mask.}
    \label{fig:system overview}
\end{figure*}

\subsection{Facial Action Coding System}
The Facial Action Coding System (FACS)  describes all visible facial movements in terms of Action Units. Combining these Action Units, one can explain virtually any facial expression \cite{FACS}. Most existing robot designs follow the FACS to replicate human-like facial expressions\cite{yan2024facial}. The authors of FACS identified six universal emotional expressions: anger, disgust, fear, happiness, sadness, and surprise\cite{expressions}. The robot design presented in this paper considers these six emotions as basic emotional expressions and tries to imitate them. 

\subsection{Humanoid Heads}

Out of various robot head types, android-type heads are the most complex and capable of producing the most realistic facial expressions. Modern high-end designs such as Ameca, Mesmer, and Aria use advanced actuation and skin-like materials to mimic human facial motion with impressive fidelity. The Desktop version of Ameca has 32-DoF solely for facial and neck movements. Aria has swappable faces with different personalities. However, their complexity and cost limit their practicality for receptionist applications.
On the other hand, there are some simpler designs in recent literature. Columbia University's EVA\cite{EVA} features 25-DoF and employs Bowden cables for actuation. It only supports symmetric facial expressions and lacks gaze control and verbal communication. Another design from Columbia, EMO\cite{EMO}, incorporates 26-DoF with rigid linkage mechanisms, magnets, and silicone skin. However, EMO also lacks verbal communication capabilities. 
Similarly, Zhibin Yan's humanoid head\cite{yan2024facial} uses a rigid linkage drive with snap button connections and silicone skin. While snap buttons ensure secure mechanical connections, they compromise the natural appearance of the design. These designs also lack human re-identification capabilities, a feature that is potentially disadvantageous in a receptionist role.

\subsection{Natural Conversation}
In our evaluation, we compared several open-source LLMs such as LLaMA 3 \cite{MetaLLama2024}, LLaMA 2, Falcon \cite{falcon}, Mixtral \cite{mixtral}, and MPT on four criteria: language understanding, generation quality, latency, and computational overhead. The LLaMA 3.1 70B, 405B and Falcon-180B variants achieved the highest benchmark scores, with MMLU above 80\% and MT-Bench close to 9, but they require substantial memory and multi-GPU resources, limiting their feasibility for low latency edge deployment.

We selected the 8 Billion parameter version of LLaMA 3.1, as it offers strong conversational performance while remaining lightweight enough for deployment on edge systems with limited computational resources. LLaMA 3’s robust benchmark results, extensive pretraining, and overall scalability made it a good choice for our robotic receptionist’s conversational engine. 

\subsection{Face Detection and Recognition}
RetinaFace\cite{retinaface} achieves high accuracy in face detection but often needs significant computational resources, making it less suitable for edge deployment. YOLO-based detectors\cite{yolov8} deliver competitive results but struggle with scalability. SCRFD\cite{SCRFD} stands out by balancing high accuracy on the WIDER Face Hard benchmark (99.7\%) with low computational requirements (7.1 GFLOPs) and fast CPU inference speeds (23ms). Its design enables efficient deployment on edge platforms while maintaining robustness across a range of face sizes. 

Regarding the face tracking, SORT\cite{SORT} offers high-speed processing but frequently suffers from identity switches due to its reliance solely on Kalman filtering.  DeepSORT\cite{DeepSORT} addresses this issue by incorporating appearance embeddings, which reduce identity switches by about 60\%, at the cost of increased computational demand. BYTETrack\cite{zhang2022bytetrack} achieved strong tracking accuracy (77.3 IDF1) and real-time CPU performance (112 FPS).

For the face recognition models, GhostFaceNets\cite{Ghost} achieves state-of-the-art accuracy on LFW (99.87\%)\cite{FaceRec} while remaining computationally efficient. AdaFace\cite{AdaFace} improves recognition performance for low-quality faces but requires large-scale training data. DiscFace\cite{DiscFace} enhances feature discrimination at the cost of increased model complexity. Despite the emergence of newer models, ArcFace\cite{arcface} remains a popular choice for real-time systems due to its strong accuracy (99.83\% on LFW)\cite{FaceRec}, stable training, and consistent inference performance under operational constraints.

%% file: sections/Methodology.tex
\subsection{System Overview}
The overview of the proposed system is illustrated in Fig.~\ref{fig:system overview}. The system consists of three main modules: the controller module, the face recognition module, and the natural conversation module. The articulated head comprises four mechanisms: eyes, eyebrows, the mouth, and the neck attached to the base. A removable silicone mask encloses the mechanical head. The Nvidia Jetson AGX Xavier serves as the central processor, managing the high-level control system of the humanoid head and running the real-time face identification model. The web camera and the 4-channel microphone array function as the system's primary visual and audio inputs. An external server handles the real-time natural language processing and audio generation algorithms required for natural conversation tasks. The subsequent sections of the paper provide detailed explanations of the system's submodules.

\subsection{Hardware Design and Implementation}
FACS-based facial simulations in MAYA 2025 were used to identify the most significant control points for facial expression generation, primarily around the lips, eyes, eyelids, and eyebrows. Based on this analysis, we designed a modular robot head with 10 control points organized into four subsystems.
\begin{figure}[!t]
\vspace{2pt}
\centering

\begin{subfigure}{0.24\columnwidth}
\centering
\includegraphics[width=\linewidth]{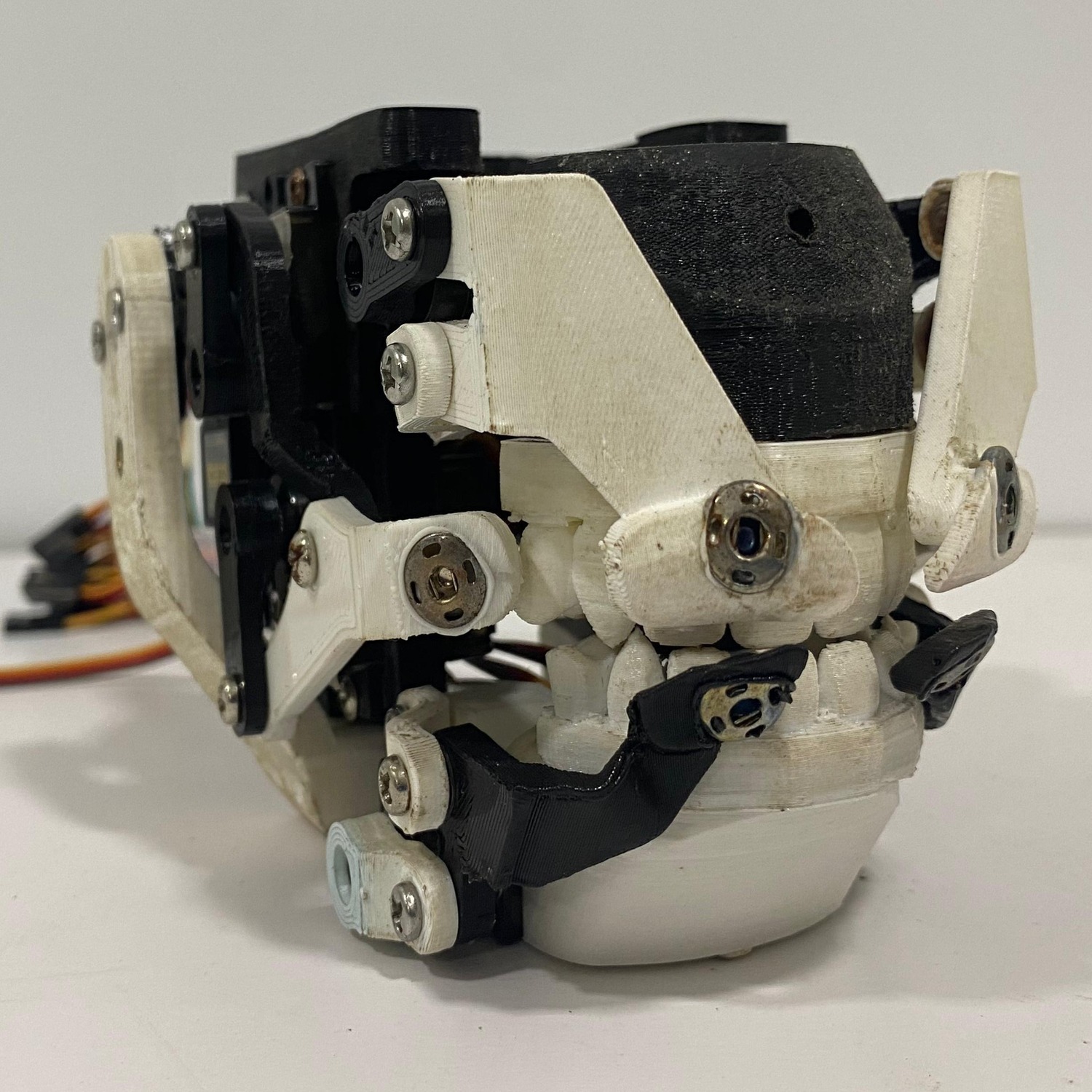}
\caption{}
\label{fig:mouth}
\end{subfigure}
\hfill
\begin{subfigure}{0.24\columnwidth}
\centering
\includegraphics[width=\linewidth]{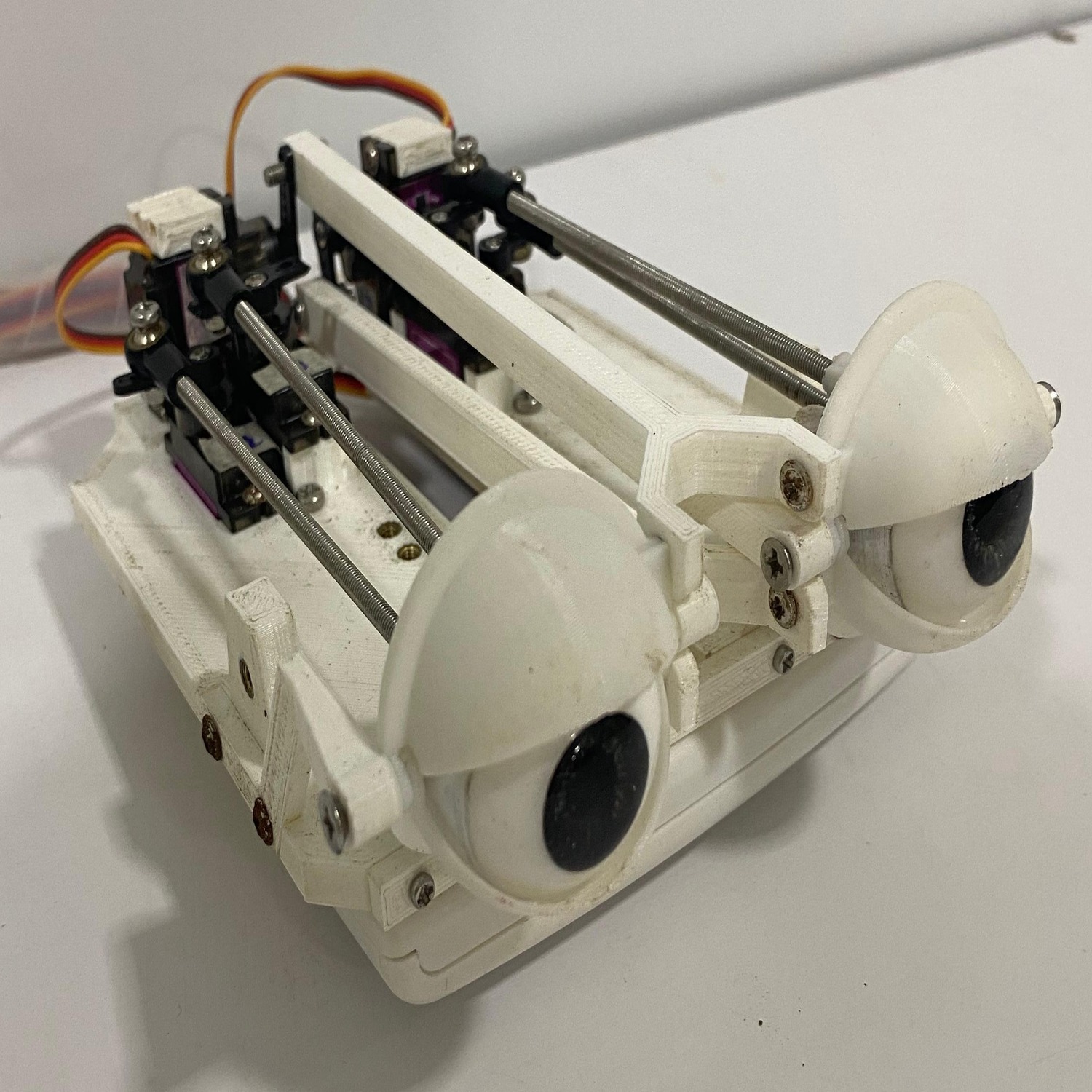}
\caption{}
\label{fig:eye}
\end{subfigure}
\hfill
\begin{subfigure}{0.24\columnwidth}
\centering
\includegraphics[width=\linewidth]{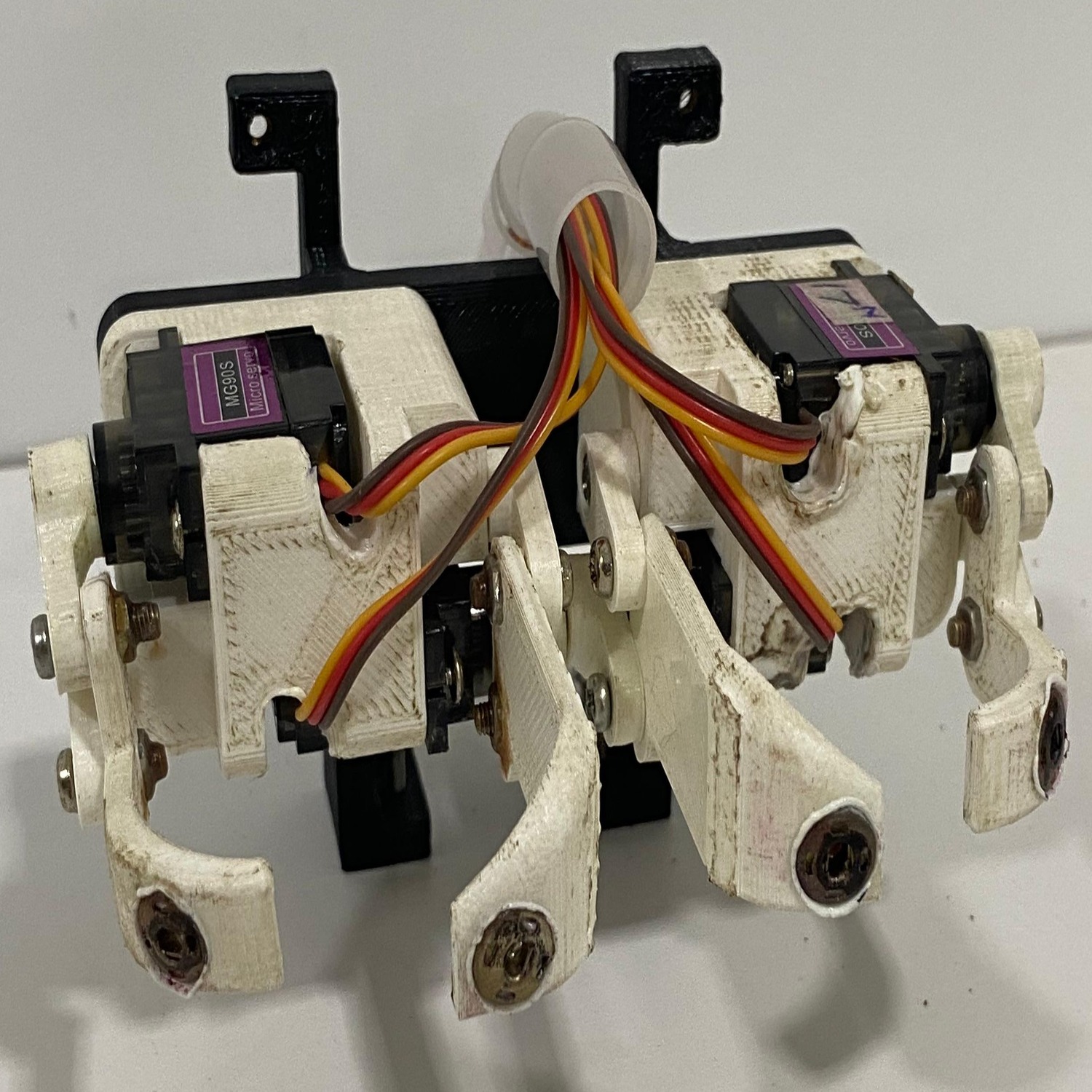}
\caption{}
\label{fig:eyebrow}
\end{subfigure}
\hfill
\begin{subfigure}{0.24\columnwidth}
\centering
\includegraphics[width=\linewidth]{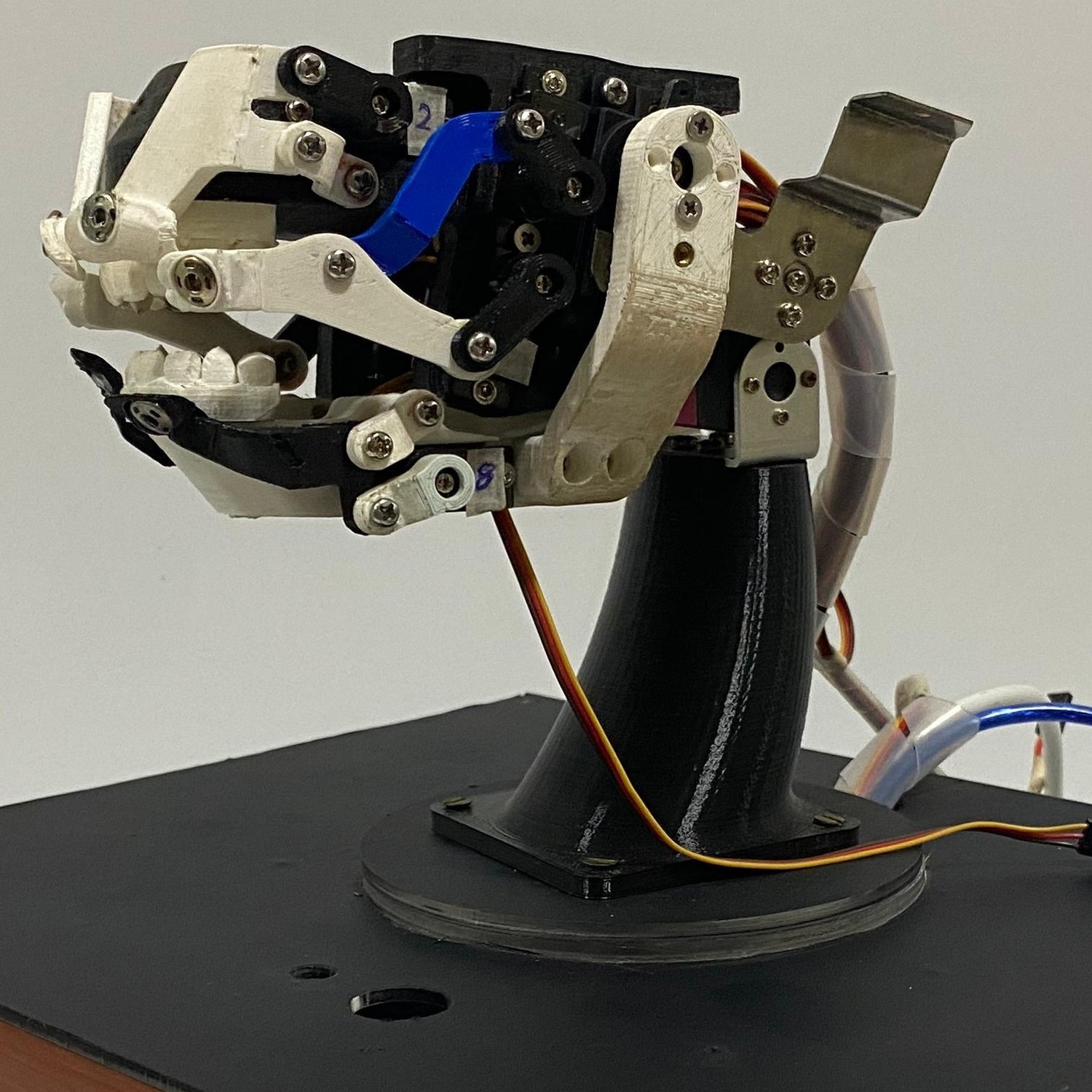}
\caption{}
\label{fig:neck}
\end{subfigure}

\caption{Key components and mechanisms of the humanoid head design. (a) Mouth mechanism, (b) eye mechanism, (c) eyebrow mechanism, and (d) neck mechanism with the base.}
\label{fig:meachanical-parts}
\end{figure}

\subsubsection{Mouth Mechanism}
The mouth mechanism has 9-DoF. Eight servo motors actuate the lips, and two redundant motors control the jaw. The four-bar linkage mechanism attached to the mouth corners enables their movement within a vertical plane. Independent motors attached to the upper and lower lips facilitate asymmetric facial expressions. 

\subsubsection{Eye Mechanism}
The humanoid head features a 6-DoF eye mechanism. Both eyeballs operate independently with 2-DoF with pitch and yaw control. 2 servo motors synchronously control the upper and lower eyelids of both eyes. Fig.~\ref{fig:eye} shows the design of the eye mechanism.

\subsubsection{Eyebrow Mechanism}
The eyebrows of the head consist of a 4-DoF mechanism with each eyebrow having 2-DoF, which independently control inner and outer brows. Fig.~\ref{fig:eyebrow} depicts the eyebrow mechanism's design.

\subsubsection{Rigid Skull and Neck Mechanism}
The neck mechanism connects the robot head to the base. With its 2-DoF, the neck moves the head to complement facial expressions such as nodding and turning. A rigid skull partially encloses the mechanical structure and supports the synthetic skin. Fig.~\ref{fig:system overview}(c) shows the exploded view of the mechanical assembly with the rigid skull. Together, the mechanical structure features a 21-DoF system powered by 22 servomotors. 

\begin{figure*}[!t]
\vspace{2pt}
    \centering
    \includegraphics[width=\linewidth]{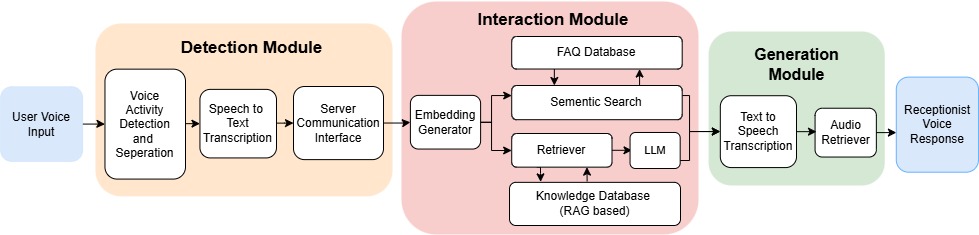}
    \caption{The overall architecture of the robot's natural conversation system: The Detection Module captures and transcribes user speech, the Interaction Module generates a relevant response, and the Generation Module delivers it as synchronized speech.}
    \label{fig:natural conversation system overview}
\end{figure*}

\subsubsection{Silicone Skin}

The silicone skin imparts a realistic, human-like appearance to the robot’s head. We modified the mold design published with the EVA robot head \cite{EVA} to fabricate the skin for our prototype. 

To the best of our knowledge, this work introduces a hybrid attachment system that integrates snap fasteners and magnets to secure the silicone skin to the rigid support structure. Previous designs employed these methods individually \cite{EMO,yan2024facial}, whereas the proposed hybrid system leverages both. Snap fasteners secure the skin to actuators at points of deformation, ensuring stable attachment during motion, while magnets hold the skin at rigid areas, allowing for straightforward attachment and removal. 

\subsection{Natural Conversation System}

Fig.~\ref{fig:natural conversation system overview} presents the overall architecture of the robot's natural conversation system, which consists of three modules: Detection, Interaction, and Generation.

\subsubsection{Detection Module}

This module utilizes inbuilt algorithm in the ReSpeaker 4-channel microphone array to detect human voice activities in 30 ms audio chunks. When voice activity is detected, recording begins immediately. If 25 consecutive chunks contain no speech, the system interprets this as a pause and sends the partial recording for speech-to-text processing, allowing early processing. End-of-speech is determined after 40 consecutive silent chunks. These thresholds were determined experimentally to balance responsiveness and accuracy.

Simultaneously, the system calculates the direction of the sound source to localize the speaker. For single speakers, it uses the built-in sound source localization algorithm in the mic array. In an environment with multiple speakers, the robot uses Open Embedded Audition System (ODAS) \cite{odas} for multiple sound sources localization, tracking, and speech separation. The recorded audio is transcribed using FasterWhisper \cite{faster-whisper}, an optimized implementation of OpenAI’s Whisper model designed for low-latency speech-to-text conversion.

\subsubsection{Interaction Module}     
The Interaction Module generates contextually accurate and domain-specific responses for user inquiries. The module converts the transcribed text into a query embedding using a pretrained sentence-transformer model, which is then processed by two parallel methods.
The first method compares the query vector with other vector embeddings in a predefined database of Frequently Asked Questions (FAQs). The FAQ database consists of commonly asked questions and their corresponding answers. The module performs a semantic search to calculate the cosine similarity. If the similarity exceeds a threshold, the system retrieves the best corresponding answer, enabling fast responses to common queries.

For unmatched queries, responses are generated using the Llama 3.1 8B Instruct model \cite{MetaLLama2024}. To maintain domain-specific responses, the system uses several methods, including a Retrieval-Augmented Generation (RAG) memory system \cite{RAG} and efficient prompt engineering techniques.

\begin{figure*}[!t]
\vspace{2pt}
    \centering
    \includegraphics[width=1\linewidth]{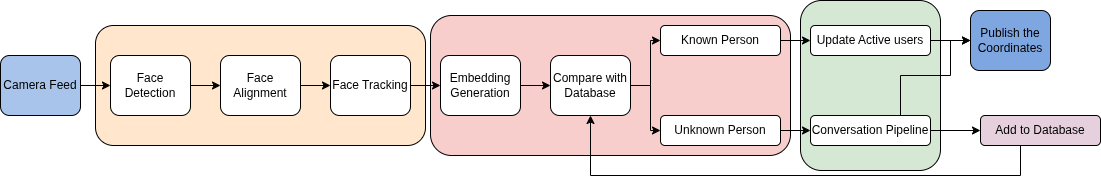}
    \caption{The overall architecture of the robot’s face recognition system: Faces are detected, aligned, and tracked, embeddings are compared with the database to identify known or unknown persons, and the system updates users, manages conversations, and publishes coordinates.}
    \label{fig:flow chart}
\end{figure*}

The RAG system includes a knowledge base stored in ChromaDB \cite{chromodb},containing department information such as staff, degree programs, research, and locations. Retrieved information and the user query are inserted into prompt templates designed for different user groups: undergraduate students, department lecturers, and visitors. Each template provides instructions for generating responses appropriate to the target audience. This approach reduces hallucinations and improves response accuracy.

\subsubsection{Generation Module}
The Generation Module delivers the robot's responses in auditory form. After the Interaction Module generates the response text, the Generation Module converts it into speech using a pre-trained MeloTTS Text-to-Speech (TTS) \cite{melotts} model. The generated audio is streamed through the robot's speakers to the user.

To synchronize voice output with mouth movements, the system analyzes the audio waveform in 10 ms chunks.The average amplitude is mapped to the jaw motor position to reflect speech intensity. A ROS2 service is then used to synchronize the motor commands with the audio playback, enabling synchronized jaw motion with speech.

\subsection{Human Re-Identification}

The proposed real-time face recognition architecture consists of four modules: Face Detection, Face Tracking, Face Alignment, and Face Recognition. Figure~\ref{fig:flow chart} shows the overall system pipeline. Each module is optimized for real-time performance and integrated using ROS2.

\subsubsection{Face Detection}

The face detection module identifies and localizes faces from the robot’s onboard camera stream. We employ SCRFD [32]. The model is executed in ONNX format using onnxruntime in the Jetson Xavier AGX. SCRFD provides bounding boxes and five-point facial landmarks for each detected face. A confidence threshold of 0.45 was selected and detections beyond 135 cm from the robot are filtered out. To address issues such as inconsistent detection under low lighting or occlusion, the module was optimized with non- maximum suppression (NMS), buffer flushing, and multi- threaded execution.

\subsubsection{Face Tracking}

BYTETrack [35] is used for face tracking. One key limitation observed was identity loss during brief occlusions or sharp head turns. To address this, a grace timeout mechanism was used so that if a previously recognized user disappears temporarily, their identity is retained for up to 3 seconds. If they reappear within this window, re-identification is avoided.

\subsubsection{Face Alignment}

Using the five-point landmarks from the detector, we apply a similarity transformation to align the face to a canonical template. The aligned output is then resized to $112 \times 112$ pixels. While ArcFace can process raw face crops, alignment significantly improves performance, especially for faces with oblique poses or partial occlusions.

\subsubsection{Face Recognition}

We use ArcFace\cite{arcface} with the iResNet100 backbone to generate 512-dimensional embedding vectors from aligned face images. These embeddings are compared to a stored database using cosine similarity. A similarity threshold of 0.4 was selected through validation experiments.

To optimize runtime performance, recognition is skipped for stable tracks where the face bounding box and orientation remain consistent across frames. Recognized individuals are labeled with either their known identity or a generated tag (\texttt{Unknown}), and their position in the frame is used to estimate gaze angles for downstream modules such as the robot's eye gaze and conversation system.

The final output is a list of currently visible users with identity labels and metadata, which is published over ROS2 for access by other nodes in the robot system.

%% file: sections/Results.tex
\begin{figure}[!t]
\vspace{2pt}
    \centering
    
    \begin{subfigure}[b]{0.15\linewidth}
        \includegraphics[width=\textwidth]{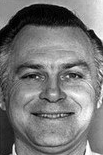}
        \caption{}
    \end{subfigure}
    \begin{subfigure}[b]{0.15\linewidth}
        \includegraphics[width=\textwidth]{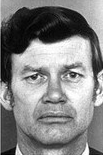}
        \caption{}
    \end{subfigure}
    \begin{subfigure}[b]{0.15\linewidth}
        \includegraphics[width=\textwidth]{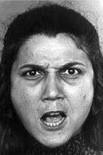}
        \caption{}
    \end{subfigure}
    \begin{subfigure}[b]{0.15\linewidth}
        \includegraphics[width=\textwidth]{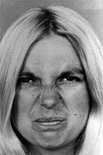}
        \caption{}
    \end{subfigure}
    \begin{subfigure}[b]{0.15\linewidth}
        \includegraphics[width=\textwidth]{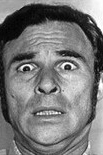}
        \caption{}
    \end{subfigure}
    \begin{subfigure}[b]{0.15\linewidth}
        \includegraphics[width=\textwidth]{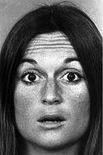}
        \caption{}
    \end{subfigure}
    
    
    \begin{subfigure}[b]{0.15\linewidth}
        \includegraphics[width=\textwidth]{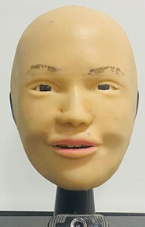}
        \caption{}
    \end{subfigure}
    \begin{subfigure}[b]{0.15\linewidth}
        \includegraphics[width=\textwidth]{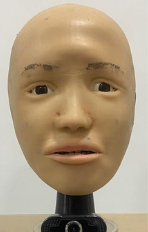}
        \caption{}
    \end{subfigure}
    \begin{subfigure}[b]{0.15\linewidth}
        \includegraphics[width=\textwidth]{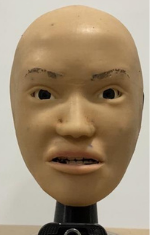}
        \caption{}
    \end{subfigure}
    \begin{subfigure}[b]{0.15\linewidth}
        \includegraphics[width=\textwidth]{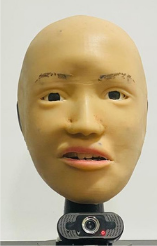}
        \caption{}
    \end{subfigure}
    \begin{subfigure}[b]{0.15\linewidth}
        \includegraphics[width=\textwidth]{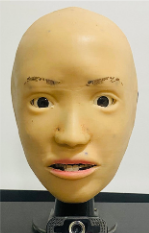}
        \caption{}
    \end{subfigure}
    \begin{subfigure}[b]{0.15\linewidth}
        \includegraphics[width=\textwidth]{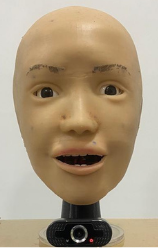}
        \caption{}
    \end{subfigure} 
    \caption{Comparison between Ekman’s six basic emotions\cite{FACS} (a–f) and the corresponding expressions generated by the humanoid robotic head (g–l). The robot mimics human-like facial expressions for happiness, sadness, anger, disgust, fear, and surprise.}
    \label{fig:facial_expressions}
\end{figure}

The Fig.~\ref{fig:facial_expressions} shows instances of the robot expressing six basic emotions. To evaluate the robot’s ability to express emotions, we conducted a user survey with 60 participants aged 20–25. Participants were shown unlabeled images of the robot displaying six basic emotions and a neutral expression, and asked to identify the emotion and rate its human-likeness on a scale from 0 (highly unlike) to 5 (highly human-like). Yan et al.\cite{yan2024facial} adopted a similar approach to evaluate user perception of their design. 

\begin{table*}[t]
\vspace{4pt}
\centering
\caption{Facial expression recognition results for the humanoid robot head.}
\begin{tabular}{c*{8}{c}cc}
\hline
\textbf{Expressed } & \multicolumn{8}{c}{\textbf{Recognized Emotion}} & \textbf{Recognition} & \textbf{Avg. Rank of} \\ 
\textbf{Emotion} & Happiness & Sadness & Anger & Surprise & Disgust & Fear & Neutral& Not identified & \textbf{Accuracy} & \textbf{Human Likeliness}\\ \hline
Happiness & 49 & 0  & 0  & 2  & 0  & 0  & 7  & 2  & 81.7\% & 4.033 \\ 
Sadness   & 0  & 48 & 0  & 1  & 1  & 10 & 0  & 0  & 80.0\% & 4.30 \\ 
Anger     & 0  & 0  & 55 & 1  & 4  & 0  & 0  & 0  & 91.7\% & 4.35 \\ 
Surprise  & 2  & 0  & 0  & 57 & 0  & 1  & 0  & 0  & 95.0\% & 4.267\\ 
Disgust   & 0  & 1  & 0  & 1  & 52 & 1  & 0  & 5  & 86.7\% & 4.25 \\ 
Fear      & 0  & 1  & 0  & 7  & 4  & 42 & 2  & 4 & 70.0\% & 3.683 \\ 
Neutral   & 6  & 0  & 0  & 0  & 0  & 0  & 50 & 4  & 83.3\% & 4.033 \\ 
\bottomrule
\end{tabular}%
\label{tab:expression recognition}
\end{table*}

Table~\ref{tab:expression recognition} summarizes the survey results for each emotion. As shown, the probability of correctly recognizing basic emotions, except for fear, exceeds 80\%. The surprise expression achieved the highest recognition rate at 95.0\%, while the fear expression had the lowest recognition rate at 70.0\%. Participants have rated the human-likeness of emotion expressions as detailed in the last column of Table~\ref{tab:expression recognition}. Since this robot is particularly designed as a receptionist, the most commonly used expressions will be neutral, happy, and surprised, which show high success rates. 
Usually, a recognition rate over 80 \% indicates that the facial expressions are easily recognizable to humans`\cite{yan2024facial}. The result of this study revealed that the proposed robot hardware is capable of expressing basic emotions except fear with a high level of anthropomorphism. 

We also evaluated key aspects of the robot’s conversational performance, focusing on response time, real-time sound source localization \cite{sound_localization_p1}, and the word error rate (WER) of the speech-to-text transcription. These metrics reflect the system’s efficiency in real-world interactions and highlight both its strengths and areas for improvement.

End-to-end response time depends on whether the response is retrieved from the FAQ database or generated by the LLM. For FAQ-based queries, the average response time was 3.14 seconds, while LLM-generated responses required 4.58 seconds due to sequential data retrieval and LLM inference in the RAG pipeline. Occasional hallucinations were observed for highly specific or domain-irrelevant queries but were largely mitigated by the RAG mechanism and prompt design.

Speech recognition was evaluated with five participants with different English accents reading a scripted passage. The system achieved an average Word Error Rate (WER) of 11.43\%. For sound source localization, a speaker was positioned at 19 angles between –90° and +90° in 10° intervals, resulting in an average angular error of 1.79°.

The face recognition system was evaluated for latency, accuracy, and multi-user performance. Using timestamps across the pipeline, the system achieved an average FPS of 12.3. Detection latency, defined as the time from frame capture to returning face bounding boxes from the SCRFD model, averaged 0.8 s. Recognition latency, which includes face alignment, ArcFace feature extraction, and cosine similarity matching with the database, averaged 1.9 s across multiple users and conditions.

For multi-user evaluation, up to four individuals were placed at different positions and angles within the camera’s field of view. The system successfully detected, tracked, and maintained consistent identities for all individuals. Using ground truth annotations, it achieved an average recognition accuracy of 89.7\%. Most errors occurred during rapid movement, occlusions, passing overs, or poor lighting, though the tracker generally reassigned consistent IDs once faces became clearly visible again.

%% file: sections/Conclusion.tex
This paper presents the design and implementation of an articulated humanoid head for a robot receptionist capable of natural human interaction. The proposed system balances mechanical simplicity and interaction capabilities by addressing limitations in existing humanoid robot heads such as high costs, complexity, and limited expressiveness. The 21-DoF structure with realistic silicone skin replicates human-like expressions and facilitates interaction. Evaluation results show high accuracy in emotion recognition and positive user feedback on conversational abilities. While the system is effective in controlled environments, improvements are needed to handle scenarios such as multi-speaker interactions. Overall, the system shows promise for receptionist applications, with future work focused on improving performance in complex scenarios. In future work, this humanoid head will be integrated with a mobile torso and articulated arms to create a complete expressive mobile robot receptionist.

%% file: ref.bib
@article{EMO,
author = {Yuhang Hu  and Boyuan Chen  and Jiong Lin  and Yunzhe Wang  and Yingke Wang  and Cameron Mehlman  and Hod Lipson },
title = {Human-robot facial coexpression},
journal = {Science Robotics},
volume = {9},
number = {88},
pages = {eadi4724},
year = {2024},
doi = {10.1126/scirobotics.adi4724},
}

@article{EVA,
title = {Facially expressive humanoid robotic face},
journal = {HardwareX},
volume = {9},
pages = {e00117},
year = {2021},
issn = {2468-0672},
doi = {https://doi.org/10.1016/j.ohx.2020.e00117},
author = {Zanwar Faraj and Mert Selamet and Carlos Morales and Patricio Torres and Maimuna Hossain and Boyuan Chen and Hod Lipson},
}

@article{FACS,
  title={Facial action coding system},
  author={Ekman, Paul and Friesen, Wallace V},
  journal={Environmental Psychology \& Nonverbal Behavior},
  year={1978}
}

@article{expressions,
author = {Ekman, Paul and Friesen, Wallace and O'Sullivan, Maureen and Chan, Anthony and Diacoyanni-Tarlatzis, Irene and Heider, Karl and Krause, Rainer and LeCompte, William and Pitcairn, Tom and Ricci Bitti, Pio and Scherer, Klaus and Tomita, Masatoshi and Tzavaras, Athanase},
year = {1987},
month = {10},
pages = {712-7},
title = {Universals and Cultural Differences in the Judgments of Facial Expressions of Emotion},
volume = {53},
journal = {Journal of personality and social psychology},
doi = {10.1037/0022-3514.53.4.712}
}

@article{yan2024facial,
  title={Facial Expression Realization of Humanoid Robot Head and Strain-Based Anthropomorphic Evaluation of Robot Facial Expressions},
  author={Yan, Zhibin and Song, Yi and Zhou, Rui and Wang, Liuwei and Wang, Zhiliang and Dai, Zhendong},
  journal={Biomimetics},
  volume={9},
  number={3},
  pages={122},
  year={2024},
  publisher={MDPI}
}

@misc{MetaLLama2024,
      title={The Llama 3 Herd of Models}, 
      author={Aaron Grattafiori and Abhimanyu Dubey and Abhinav Jauhri and Abhinav Pandey and Abhishek Kadian and Ahmad Al-Dahle},
      year={2024},
      eprint={2407.21783},
      archivePrefix={arXiv},
      primaryClass={cs.AI},
      url={https://arxiv.org/abs/2407.21783}, 
}

@misc{faster-whisper,
      title={Robust Speech Recognition via Large-Scale Weak Supervision}, 
      author={Alec Radford and Jong Wook Kim and Tao Xu and Greg Brockman and Christine McLeavey and Ilya Sutskever},
      year={2022},
      eprint={2212.04356},
      archivePrefix={arXiv},
      primaryClass={eess.AS},
      url={https://arxiv.org/abs/2212.04356}, 
}

@techreport{chromodb,
  title = {Evaluating Chunking Strategies for Retrieval},
  author = {Smith, Brandon and Troynikov, Anton},
  year = {2024},
  month = {July},
  institution = {Chroma},
  url = {https://research.trychroma.com/evaluating-chunking},
}

@ARTICLE{odas,
    AUTHOR={Grondin, François  and Létourneau, Dominic  and Godin, Cédric  and Lauzon, Jean-Samuel  and Vincent, Jonathan  and Michaud, Simon  and Faucher, Samuel  and Michaud, François },
    TITLE={ODAS: Open embeddeD Audition System},
    JOURNAL={Frontiers in Robotics and AI},
    VOLUME={Volume 9 - 2022},
    YEAR={2022},
    URL={https://www.frontiersin.org/journals/robotics-and-ai/articles/10.3389/frobt.2022.854444},
    DOI={10.3389/frobt.2022.854444},
    ISSN={2296-9144}
}

@software{melotts,
  author={Zhao, Wenliang and Yu, Xumin and Qin, Zengyi},
  title = {MeloTTS: High-quality Multi-lingual Multi-accent Text-to-Speech},
  url = {https://github.com/myshell-ai/MeloTTS},
  year = {2023}
}

@ARTICLE{Ghost,
  author={Alansari, Mohamad and Hay, Oussama Abdul and Javed, Sajid and Shoufan, Abdulhadi and Zweiri, Yahya and Werghi, Naoufel},
  journal={IEEE Access}, 
  title={GhostFaceNets: Lightweight Face Recognition Model From Cheap Operations}, 
  year={2023},
  volume={11},
  number={},
  pages={35429-35446},
  doi={10.1109/ACCESS.2023.3266068}}

@INPROCEEDINGS{yolov8,
  author={Varghese, Rejin and M., Sambath},
  booktitle={2024 International Conference on Advances in Data Engineering and Intelligent Computing Systems (ADICS)}, 
  title={YOLOv8: A Novel Object Detection Algorithm with Enhanced Performance and Robustness}, 
  year={2024},
  volume={},
  number={},
  pages={1-6},
  doi={10.1109/ADICS58448.2024.10533619}}

@inproceedings{retinaface,
  title={Retinaface: Single-shot multi-level face localisation in the wild},
  author={Deng, Jiankang and Guo, Jia and Ververas, Evangelos and Kotsia, Irene and Zafeiriou, Stefanos},
  booktitle={Proceedings of the IEEE/CVF conference on computer vision and pattern recognition},
  pages={5203--5212},
  year={2020}
}

@inproceedings{SORT,
  title={Simple online and realtime tracking},
  author={Bewley, Alex and Ge, Zongyuan and Ott, Lionel and Ramos, Fabio and Upcroft, Ben},
  booktitle={2016 IEEE international conference on image processing (ICIP)},
  pages={3464--3468},
  year={2016},
  organization={Ieee}
}

@inproceedings{DeepSORT,
  title={Simple online and realtime tracking with a deep association metric},
  author={Wojke, Nicolai and Bewley, Alex and Paulus, Dietrich},
  booktitle={2017 IEEE international conference on image processing (ICIP)},
  pages={3645--3649},
  year={2017},
  organization={IEEE}
}

@inproceedings{zhang2022bytetrack,
  title={Bytetrack: Multi-object tracking by associating every detection box},
  author={Zhang, Yifu and Sun, Peize and Jiang, Yi and Yu, Dongdong and Weng, Fucheng and Yuan, Zehuan and Luo, Ping and Liu, Wenyu and Wang, Xinggang},
  booktitle={European conference on computer vision},
  pages={1--21},
  year={2022},
  organization={Springer}
}

@misc{SCRFD,
                    title={Sample and Computation Redistribution for Efficient Face Detection}, 
                    author={Jia Guo and Jiankang Deng and Alexandros Lattas and Stefanos Zafeiriou},
                    year={2021},
                    eprint={2105.04714},
                    archivePrefix={arXiv},
                    primaryClass={cs.CV}
              }

@online{FaceRec,
  title     = "{Performance on LFW Dataset}",
  author    = "{paperswithcode}",
  url       = {https://paperswithcode.com/sota/face-recognition-on-lfw},
}

@inbook{DiscFace,
author = {Kim, Insoo and Han, Seungju and Park, Seong-Jin and Baek, Ji-won and Shin, Jinwoo and Han, Jae Joon and Choi, Changkyu},
year = {2021},
month = {02},
pages = {358-374},
title = {DiscFace: Minimum Discrepancy Learning for Deep Face Recognition},
isbn = {978-3-030-69540-8},
doi = {10.1007/978-3-030-69541-5_22}
}

@INPROCEEDINGS{AdaFace,
  author={Kim, Minchul and Jain, Anil K. and Liu, Xiaoming},
  booktitle={2022 IEEE/CVF Conference on Computer Vision and Pattern Recognition (CVPR)}, 
  title={AdaFace: Quality Adaptive Margin for Face Recognition}, 
  year={2022},
  volume={},
  number={},
  pages={18729-18738},
  doi={10.1109/CVPR52688.2022.01819}}

@inproceedings{arcface,
  title={Arcface: Additive angular margin loss for deep face recognition},
  author={Deng, Jiankang and Guo, Jia and Xue, Niannan and Zafeiriou, Stefanos},
  booktitle={Proceedings of the IEEE/CVF conference on computer vision and pattern recognition},
  pages={4690--4699},
  year={2019}
}

@article{falcon,
  title={The Falcon Series of Open Language Models},
  author={Ebtesam Almazrouei and Hamza Alobeidli and Abdulaziz Alshamsi and Alessandro Cappelli and Ruxandra-Aim{\'e}e Cojocaru and Daniel Hesslow and Julien Launay and Quentin Malartic and Daniele Mazzotta and Badreddine Noune and Baptiste Pannier and Guilherme Penedo},
  journal={ArXiv},
  year={2023},
  volume={abs/2311.16867},
  url={https://api.semanticscholar.org/CorpusID:265466629}
}

@article{mixtral,
  title={Mixtral of Experts},
  author={Albert Q. Jiang and Alexandre Sablayrolles and Antoine Roux and Arthur Mensch and Blanche Savary and Chris Bamford and Devendra Singh Chaplot},
  journal={ArXiv},
  year={2024},
  volume={abs/2401.04088},
  url={https://api.semanticscholar.org/CorpusID:266844877}
}

@article{RAG,
  title={Retrieval-Augmented Generation for Knowledge-Intensive NLP Tasks},
  author={Patrick Lewis and Ethan Perez and Aleksandara Piktus and Fabio Petroni and Vladimir Karpukhin and Naman Goyal and Heinrich Kuttler and Mike Lewis and Wen-tau Yih and Tim Rockt{\"a}schel and Sebastian Riedel and Douwe Kiela},
  journal={ArXiv},
  year={2020},
  volume={abs/2005.11401},
  url={https://api.semanticscholar.org/CorpusID:218869575}
}

@Article{sound_localization_p1,
AUTHOR = {Shi, Zhanbo and Zhang, Lin and Wang, Dongqing},
TITLE = {Audio–Visual Sound Source Localization and Tracking Based on Mobile Robot for The Cocktail Party Problem},
JOURNAL = {Applied Sciences},
VOLUME = {13},
YEAR = {2023},
NUMBER = {10},
ARTICLE-NUMBER = {6056},
URL = {https://www.mdpi.com/2076-3417/13/10/6056},
ISSN = {2076-3417},
DOI = {10.3390/app13106056}
}

@misc{EngineeredArts2025,
  title        = {Engineered Arts — Embodied AI Social Humanoid Robots},
  author       = {{Engineered Arts}},
  year         = {2025},
  note         = {Accessed: 2025-10-01},
  url          = {https://engineeredarts.com/}
}

@misc{Realbotix2025,
  title        = {Realbotix — Robots for Human Interaction},
  author       = {{Realbotix}},
  year         = {2025},
  note         = {Accessed: 2025-10-01},
  url          = {https://www.realbotix.com/}
}

@misc{mesmer,
  author       = {{Engineered Arts}},
  title        = {{Mesmer – Social Humanoid Robot}},
  howpublished = {\url{https://engineeredarts.com/robots/mesmer}},
  note         = {Accessed: Mar. 11, 2026}
}
